\documentclass[acmsmall, anonymous=false, nonacm=true]{acmart}

\usepackage{verbatim}
\usepackage{mdframed}
\usepackage{amsmath}
\usepackage{xcolor}
\usepackage{bm} 
\usepackage{csquotes}
\usepackage{float}
\usepackage{subcaption}
\usepackage{wrapfig}
\usepackage{tikz}
\usepackage{listings}
\usepackage{graphbox}
\usepackage{xspace}
\usepackage{multicol}
\usepackage{relsize}
\usepackage{changepage}
\usetikzlibrary{shapes,arrows}

\mdfdefinestyle{examplebox}{
    splittopskip=0,%
    splitbottomskip=0,%
    frametitleaboveskip=0,
    frametitlebelowskip=0,
    skipabove=0,%
    skipbelow=0,%
    innerleftmargin=1mm,%
    innerrightmargin=1mm,%
    innertopmargin=1mm,%
    innerbottommargin=1mm,%
    roundcorner=0mm}

\AtBeginDocument{%
  \providecommand\BibTeX{{%
    \normalfont B\kern-0.5em{\scshape i\kern-0.25em b}\kern-0.8em\TeX}}}

\begin{document}

\title{ADiag: Graph Neural Network Based Diagnosis of Alzheimer’s Disease}


\author{Vishnu Ram Sampathkumar}
\affiliation{%
  \institution{Department of Neuroimaging and Interventional Radiology, National Institute of Mental Health and Neurosciences (NIMHANS)}
  \country{Bengaluru, 560029}
}


\renewcommand{\shortauthors}{V. R. Sampathkumar}

\begin{abstract}
Alzheimer’s Disease (AD) is the most widespread neurodegenerative disease, affecting over 50
million people across the world. While its progression cannot be stopped, early and accurate
diagnostic testing can drastically improve quality of life in patients. Currently, only qualitative means
of testing are employed in the form of scoring performance on a battery of cognitive tests. The
inherent disadvantage of this method is that the burden of an accurate diagnosis falls on the clinician’s
competence. Quantitative methods like MRI scan assessment are inaccurate at best, due to the elusive
nature of visually observable changes in the brain. In lieu of these disadvantages to extant methods of
AD diagnosis, we have developed ADiag, a novel quantitative method to diagnose AD through
GraphSAGE Network and Dense Differentiable Pooling (DDP) analysis of large graphs based on
thickness difference between different structural regions of the cortex. Preliminary tests of ADiag
have revealed a robust accuracy of 83\%, vastly outperforming other qualitative and quantitative diagnostic techniques.

\end{abstract}

\maketitle

\section{Introduction}
\label{s:intro}
\subsection{Overview: Alzheimer's Disease}
Alzheimer’s Disease (AD) is a progressive neurodegenerative disease affecting more than 50 million people worldwide \cite{nichols2019global}. Most patients are above the age of 65, though early onset forms do exist. AD is initially characterized by mild confusion and forgetfulness (known clinically as Mild Cognitive Impairment: MCI), but soon progresses into increasing memory loss and cognitive impairment; patients with stage 7 (final stage) AD lose all awareness of themselves and their surroundings, and are unable to even make basic muscle movements. Though AD is usually not the direct cause of death, it is directly linked to other terminal pathologies like pneumonia. The cause of AD is not precisely known, but the buildup of aggregations of misfolded proteins (beta-amyloid and tau proteins) in the hippocampus and the temporal lobe has been cited as a factor. These aggregations are neurotoxic and progressively kill cortical neurons; this process can be understood as a progressive decrease in cortical thickness.
\subsection{Diagnostic Methodologies}
No treatments exist yet, but timely diagnosis can contribute to an improved quality of life for the patient. These diagnostic methods, however are extremely qualitative in nature; one such test, the Mini Mental State Examination (MMSE) requires clinicians to evaluate potential patients on a 30-point test on attention, memory, language, orientation and visual-spatial skills. Another is the Clinical Dementia Rating (CDR), which  is a 5-point test on a similar rubric, but also includes home affairs and personal care \cite{balsis2015scores}. Irrespective of the specifics, a correct diagnosis is solely based on the clinician’s competence and not on quantitative backing; this is the major cause of the high rate of misdiagnosis. In  lieu of this, it is absolutely essential that efficient quantitative methods of AD diagnosis are developed.Quantitative AD testing is restricted to cerebral biopsy \cite{warren2005brain}, and is not widely implemented: being invasive, there is a high risk of infection, which can be debilitating for senior citizens. In fact, a conclusive diagnosis of AD is done only after the patient has passed away and an autopsy has been performed. Other non-invasive quantitative methods have been developed, but are still experimental; they rely almost exclusively on supervised learning, and are thus inherently data hungry and inaccurate.
Graph Neural Networks (GNNs) are a powerful tool to aid in AD diagnosis, simply because the brain can also be represented as a network graph, defined by distinct nodes (representing different structural and
functional regions of the brain), and the edges, representing neuronal
connections between these regions of the brain. ADiag is thus a GNN model that leverages the graph representation of the brain to diagnose Alzheimer's Disease.

\section{Methods}

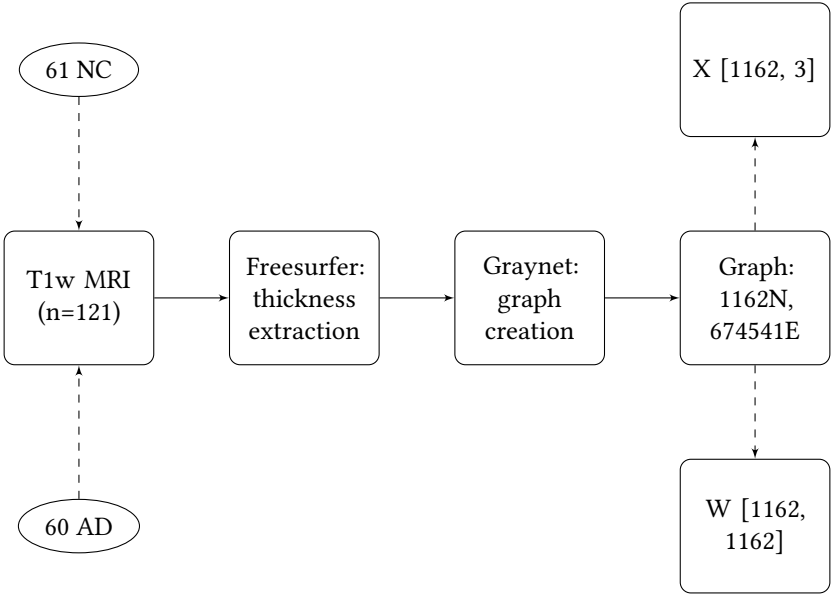
\begin{figure}[H]
    \centering
   
    \label{fig:sdgel}

\tikzstyle{decision} = [diamond, draw,  
    text width=4.5em, text badly centered, node distance=3cm, inner sep=0pt]
\tikzstyle{block} = [rectangle, draw, 
    text width=5em, text centered, rounded corners, minimum height=5em]
\tikzstyle{line} = [draw, -latex']
\tikzstyle{cloud} = [draw, ellipse, node distance=3cm,
    minimum height=2em]
    
\begin{tikzpicture}[node distance = 3cm, auto]
    \node [block] (init) {T1w MRI (n=121)};
    \node [cloud, above of=init] (NC) {61 NC};
    \node [cloud, below of=init] (AD) {60 AD};
    \node [block, right of=init] (FreeSurfer) {Freesurfer: thickness extraction};
    \node [block, right of=FreeSurfer] (evaluate) {Graynet:  graph creation};
    \node [block, right of=evaluate] (decide) {Graph: 1162N, 674541E};
    \node [block, below of=decide] (Adj) {W [1162, 1162]};
    \node [block, above of=decide] (NF) {X [1162, 3]};
    \path [line] (init) -- (FreeSurfer);
    \path [line] (FreeSurfer) -- (evaluate);
    \path [line] (evaluate) -- (decide);

    \path [line,dashed] (NC) -- (init);
    \path [line,dashed] (AD) -- (init);
    
    \path [line, dashed] (decide) -- (Adj);
    \path [line, dashed] (decide) -- (NF);

\end{tikzpicture}

 \caption{Procedural Overview of ADiag}

\end{figure}

\subsection{Data Acquisition and Pre-processing}

Data was acquired from the OASIS 3 database \cite{lamontagne2019oasis} in the form of T1 weighted MRI scans; these scans were processed on FreeSurfer to generate thickness data, and this was further processed in Graynet to generate network graphs. Overall, this was a computationally demanding stage of the project, where processing of each scan took upwards of 7 hours.

\subsubsection{Database Selection and Inclusion Criteria}

The OASIS 3 database was chosen due to the abundance of high quality T1 weighted MRI scans. Scans were subdivided into AD group (AD) and the Normal Controls group (NC); the criteria for this subdivision was the MMSE and CDR score. 121 scans were chosen in all: 60 scans belonging to AD,and 61 scans belonging to NC.

\begin{table}[H]
\centering
\begin{tabular}{ccc}
\hline
Group & CDR & MMSE \\
\hline
Normal Control (NC) & $ 0 < $ & $ 25-30 $ \\
Alzheimer's Group (AD) & $ 0.5 < $ & $ < 24 $ \\
\hline
\end{tabular}

\vspace*{5mm}

\label{tab:shape-functions}
\caption{\bf Criteria for Subdivision into AD and NC \cite{querbes2009early}}
\end{table}

\subsubsection{FreeSurfer Processing}

The scans were then processed on FreeSurfer, an open access, industry standard MRI analysis software \cite{fischl2012freesurfer}. QCache and ReconAll flags were applied during scan analysis to ensure that cortical thickness features were properly extracted; the data was then subjected to the quality control software, VisualQC to ensure that no irregularities were present in the data due to bias-field or regions of abnormal signal intensity .

\subsubsection{Graynet Processing}

The thickness data extracted by FreeSurfer was then processed by the graph generation software Graynet \cite{raamana2018graynet}; here, differences in cortical thickness between functional regions of the brain were used as parameters to define the weight of the different
connections. Large differences in thickness were taken to define relatively weaker connections than those defined by smaller thickness differences. Obviously, the strength of these connections was defined by the edge weight $e_{i,j}$ between connected nodes $i$ and $j$.
These nodes were taken to be aggregations of vertices (defined by absolute cortical thickness values); the number of vertices incorporated into each node was inversely proportional to the number of nodes in the graph, and thus the size of the graph. To preserve the detail and intricacies of the thickness disparities, we chose a relatively small number of 250 vertices to be aggregated into each node. The graph of each scan had exactly 1162 nodes, and
6,74,541 edges between these nodes.

\subsubsection{Graph Attributes}

Each graph consisted of x, the node feature matrix of
shape [1162, 3] where each node had three features,
representing the three dimensional x, y, z coordinates of
the node in the brain. The graph was also characterized
by the edge index, a sparse matrix in COO format that
listed the connections between each node; the weighted
adjacency matrix W, and the label y. These attributes of each graph defined a unique Pytorch
Geometric ‘data’ object; together, each data object
formed a Pytorch Geometric dataset.

\begin{figure}[H]
\begin{center}

\centering
\includegraphics[height=7cm]{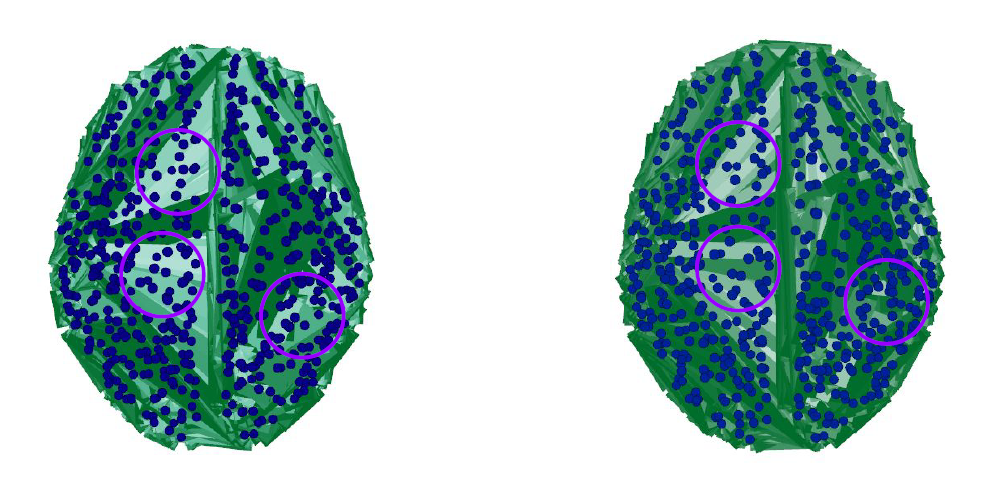}
\caption{AD graph (left) and NC graph (right) with edges (in green) ranked according to weight;
though they have a similar appearance, AG scan has noticeably lighter edges (weaker connections),
circled in purple}
\label{fig:fc1}
\end{center}
\newpage
\end{figure}

\subsection{Neural Network Classification}

Our model was constructed using Pytorch 1.6.0 and
Pytorch Geometric 1.6.1 \cite{paszke2019pytorch} on an NVIDIA RTX 2060S
with CUDA 11.0 compatibility.

\subsubsection{GraphSAGE layer}
Our model consisted of three dense GraphSAGE \cite{hamilton2017inductive} 
convolutional layers that took the number of input,
output and hidden channels as input parameters.
Each of these layers performed inductive learning on the
graphs by aggregating nodal information at each
successive iteration, thereby simultaneously
incorporating information from far reaches of the graph,
and increasing the amount of information available to
each node.
This aggregation operation was performed according to
the formula: 

\begin{equation}
    \mathlarger{h_{N(v)}^{k}\leftarrow Aggregate_{k}({h_{u}^{k-1},  \forall u\in N(v)})}
\end{equation}

\begin{equation}
\mathlarger{h_{N(v)}^{k}\leftarrow \sigma (W_{k} \cdot  Concat({h_{v}^{k-1}, h_{N(v)}}))}
\end{equation}

\vspace*{4.1mm}

where,  $h_{N(v)}^{k}$ represents the node representation
vector of an aggregation of k nodes present in a single
neighborhood, $N$ , and $h_{u}^{k-1}$ represents the node
representation vector of each node in this neighborhood.
The node representation vector is then concatenated with
the neighborhood representation vector, and this is fed
through a fully connected layer having sigmoid
activation function.

The GraphSAGE layer has a number of advantages, of
which two are salient: the inductive learning process
allows for information-rich, aggregated representations
of each node, and the neighborhood aggregation process
allows for even unseen nodes to be included in the
learning process.

\subsubsection{Dense Differentiable Pooling Layer}

The processed graph outputted by the GraphSAGE layer
then undergoes pooling in the DDP layer \cite{ying2018hierarchical}; the goal of the
pooling layer is to methodically reduce the coarsen the
graph from 1162 nodes down to 18 nodes so that it can
be easily classified.
Each pooling layer performs the coarsening operation
according to the formula:

\begin{equation}
    \mathlarger{X' = softmax(S)^{T}\cdot X}
\end{equation}

\begin{equation}
    \mathlarger{A' = softmax(S)^{T}\cdot (A) \cdot softmax(S)}
\end{equation}

\vspace*{4.1mm}

Where, $X$ and $X’$ are the initial and final node feature
matrices, respectively, and $A’$ and $A$ are the initial and
final adjacency feature matrices, respectively. $S$ is the
assignment tensor, which governs the coarsening of the
graph, and is an output of the GraphSAGE layer.
Each pooling layer reduced the number of nodes by
approximately 75\%, such that the graph was coarsened
down to 18 nodes before being fed to the fully connected
layers for binary classification.

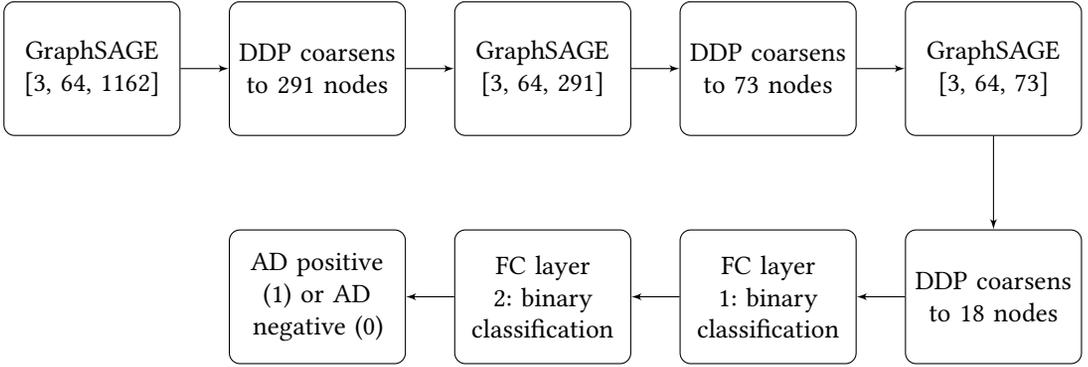
\begin{figure}[H]

    \label{fig:my_label}

\tikzstyle{decision} = [diamond, draw,  
    text width=4.5em, text badly centered, node distance=3cm, inner sep=0pt]
\tikzstyle{block} = [rectangle, draw, 
    text width=6em, text centered, rounded corners, minimum height=5em]
\tikzstyle{line} = [draw, -latex']
\tikzstyle{cloud} = [draw, ellipse, node distance=cm,
    minimum height=2em]
    
\begin{tikzpicture}[node distance = 3cm, auto]
    \node [block] (GS1) {GraphSAGE [3, 64, 1162]};

    \node [block, right of=GS1] (DD1) {DDP coarsens to 291 nodes};
    \node [block, right of=DD1] (GS2) {GraphSAGE [3, 64, 291]};
    \node [block, right of=GS2] (DD2) {DDP coarsens to 73 nodes};
    \node [block, right of=DD2] (GS3) {GraphSAGE [3, 64, 73]};
    \node [block, below of=GS3] (DD3) {DDP coarsens to 18 nodes};
    \node [block, left of=DD3] (FC1) {FC layer 1: binary classification};
    \node [block, left of=FC1] (FC2) {FC layer 2: binary classification};
    \node [block, left of=FC2] (decision) {AD positive (1) or AD negative (0)};

    \path [line] (GS1) -- (DD1);
    \path [line] (DD1) -- (GS2);
    \path [line] (GS2) -- (DD2);
    \path [line] (DD2) -- (GS3);
    \path [line] (GS3) -- (DD3);
    \path [line] (DD3) -- (FC1);
    \path [line] (FC1) -- (FC2);
    \path [line] (FC2) -- (decision);

\end{tikzpicture}
\caption{Neural Network Workflow}

\end{figure}

\section{Results}
After running the model over 150 epochs, we observed
a peak validation accuracy of 80.1\% and a training loss
of 0.71. To improve accuracy and decrease training loss,
we applied batch normalisation, and applied learning
rate optimisation. This increased accuracy up to 83.44\%,
and decreased loss to 0.695.
These accuracy and loss values, however, are
preliminary and will almost definitely be drastically
improved with increased training data. Currently, only
121 graphs were used for ADiag due to the high
processing time per graph and the resulting time
constraint. We have, however, managed to automate this
process, and expect to have at least 300 graphs forming
our dataset in the near future; this is expected to increase
accuracy values up to at least 95\% and decrease loss to
0.4. Furthermore, we are planning to develop an original
neural network architecture incorporating attention
weights expressly for ADiag. This will allow calculation
of each node’s importance with respect to the other, and
thus further empower the neighborhood-specific focus of
the model.

\begin{figure}[H]
\centering
\begin{minipage}{.5\textwidth}
  \centering
  \includegraphics[width=6cm, height=4cm]{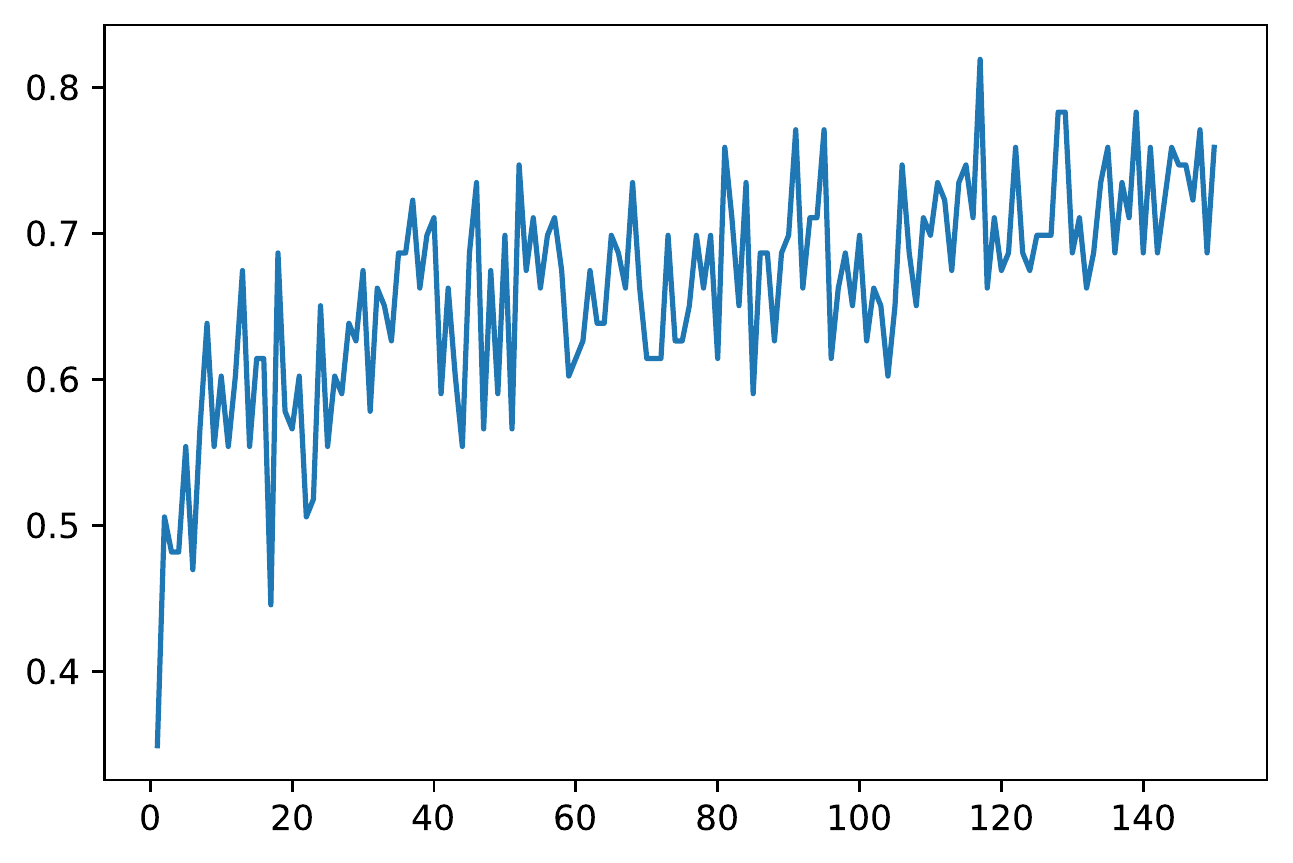}
  \captionof{figure}{Accuracy versus Epoch; peak at 83.44\%}
  \label{fig:test1}
\end{minipage}%
\begin{minipage}{.5\textwidth}
  \centering
  \includegraphics[width=6cm, height=4cm]{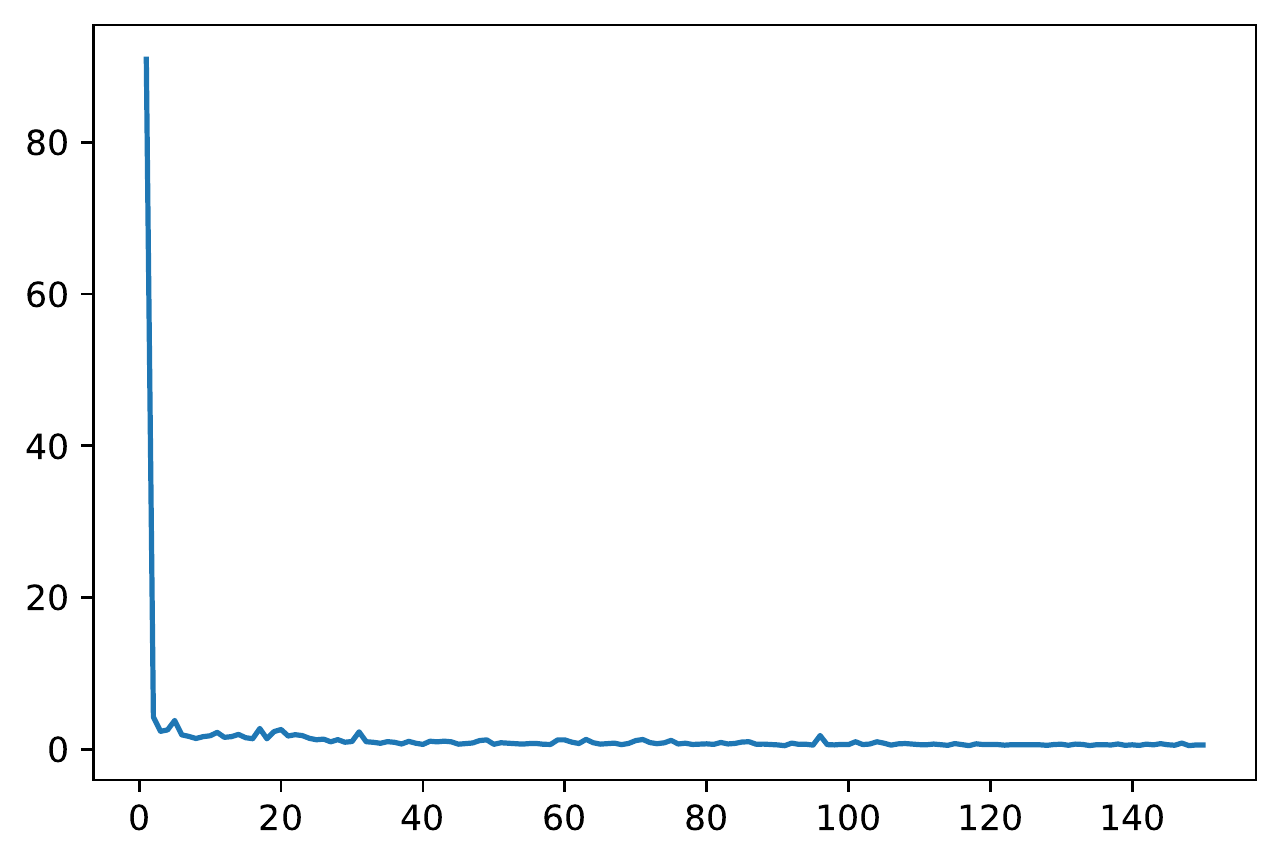}
  \captionof{figure}{Training Loss versus Epoch; trough at 0.695}
  \label{fig:test2}
\end{minipage}

\end{figure}

\section{Conclusion and Future Steps}
ADiag is superior in each and every respect compared to
extant AD diagnostic methods due to its non-invasive
nature, use of commonly available T1 weighted MRI scans,
superior accuracy, high scalability and robust
construction. Combined with an expansion in dataset
size from 121 to 300 graphs and a revamped architecture,
ADiag’s accuracy values are expected to increase to at
least 95\%. Its use of an extremely dynamic data type like
graphs means that it can be trained on data dimensionality is not consequential. ADiag is a worthy tool for clinicians to quantitatively diagnose AD; we thus hope to reduce
misdiagnosis rates and forge a path to decisive diagnosis
of AD for patients' improved quality of life.

\section{Acknowledgements}

I sincerely thank Tejas Jayashankar (PhD student, EECS department, MIT), Vrishab
Krishna (past ISEF finalist), and Siddharth Muralidaran (graduate student, ECE, UIUC) for
being my mentors and guides over the course of my research work. I also thank Dr.
Chandana Nagaraj, Associate Professor, Department of Nuclear Medicine and
Interventional Radiology, NIMHANS, for guiding me through the medical imaging aspects
of the project, and Dr. Partha Talukdar, Associate Professor, Department of
Computational and Data Sciences, IISc, for guiding me through the computational aspects
of the project. 

\bibliographystyle{IEEEtranN}
\bibliography{citations}

\end{document}